\documentclass{article}

\usepackage[utf8]{inputenc}
\usepackage{amsmath, amssymb, latexsym}
\usepackage{sidecap}
\usepackage{algorithm}
\usepackage{algorithmic}
\usepackage{biblatex}
\usepackage{tikz}
\usepackage{mathtools}
\usepackage[toc,page]{appendix}
\usepackage{hyperref}
\DeclarePairedDelimiter\ceil{\lceil}{\rceil}

\title{An Analytical Approach to Compute the Exact Preimage of Feed-Forward Neural Networks}
\author{Théo Nancy, Vassili Maillet, Johann Barbier }

\begin{document}
\maketitle

\begin{abstract}
    Neural networks are a convenient way to automatically fit functions that are too complex to be described by hand. 
    The downside of this approach is that it leads to the creation of a black-box. 
    Finding the preimage of a given output would help to better understand how and why such neural networks had given such outputs. 
    Because most neural networks are noninjective functions, it is often impossible to compute them using only a numerical way. 
    The point of this study is to give a method to compute the exact preimage of any Feed-Forward Neural Network with linear or piecewise 
    linear activation functions for hidden layers. In contrast to other methods, this one is not returning a unique solution for a unique output 
    but returns analytically the entire and exact preimage.
\end{abstract}

\newpage
\section*{Notations and assumptions}
\subsection*{Notations}
\subsubsection*{Numbers and Arrays}

 \begin{tabular}{p{2cm}p{10cm}p{1cm}}
    $a,A$   & A scalar (real or integer)\\
    $\textbf{a}$  & A vector \\
    $\textbf{A}$  & A matrix\\
    $\textbf{I}_{n}$  & The identity matrix of size $n$\\
    $\textbf{0}_{n,m}$  & The null matrix of dimensions $n \times m$
    \end{tabular}
    
\subsubsection*{Indexing}
\begin{tabular}{p{2cm}p{10cm}p{1cm}}
    $a_{i}$   & Element $i$ of a vector $\textbf{a}$, with indexing starting at 1\\
    \end{tabular}

\subsubsection*{Sets}
\begin{tabular}{p{2cm}p{10cm}p{1cm}}
            $\mathbb{A}$   & A set  \\
            $[a,b]$ & The real interval including $a$ and $b$ \\
            $[a,b[$ & The real interval including $a$ but excluding $b$ \\
            $[\![n,m]\!]$ & The integer interval including $n$ and $m$ \\
            $[\![n,m[\![$ & The integer interval including $n$ but excluding $m$ \\
            \end{tabular}

\subsubsection*{Functions}
\begin{tabular}{p{2cm}p{10cm}p{1cm}}
        $f:\mathbb{A} \longrightarrow \mathbb{B}$   & The function $f$ with domain $\mathbb{A}$ and codomain $\mathbb{B}$ \\
        $f^{-1}$ & The preimage function of $f$ \\
        $sgn$ & The sign function \\
        $Id$ & The identity function \\
        $ReLU$ & The Rectified Linear Unit function \\
        $PReLU$ & The Parametric Rectified Linear Unit function \\
        \end{tabular}

\subsubsection*{Abbreviations}
\begin{tabular}{p{2cm}p{10cm}p{1cm}}
        $ANN$   & Artificial Neural Network\\
        $FFNN$ & Feed Fordward Neural Network \\
        \end{tabular}

\subsection*{Assumptions}
The Lesbegue measure of the set of singular matrices is zero. Note that the probability for a random matrix to be singular is not zero 
on a computer because of their finite precision. However it remains very unlikely for a random squared matrix to be singular in theory. 
As a result we will assume that all squared matrices in this study are invertible.
\newpage
\tableofcontents
\newpage
\section{Introduction}
The solutions of some problems are easy to explain by giving multiple examples but tremendously hard to describe analytically. 
This is for example the case of image recognition. The difference between a dog and a cat picture is instantly discovered by a human eye. 
It is however almost impossible to mathematically explain what is a dog and what is a cat. Since AlexNet in 2012 \cite{krizhevsky2012imagenet}, 
Artificial Neural Networks (ANNs) have been shown as a wonderful way for computer scientists to solve complex machine learning problems \cite{sejnowski2020unreasonable}. 
Thanks to repeated training on examples, ANNs can learn to solve such complex problems as image recognition by fitting a large function 
that describe best the given problem depending on the given examples.\\
According to the universal approximation theorem, ReLU networks with $n$ + 1 layers can approximate any continuous function of 
$n$-dimensional input variables \cite{hanin2017approximating}. It means that every problem where the solution can be expressed as a continuous function can 
theoretically be  solved by an ANN. \\
If the ANN correctly approximates the function, most of the information about how the problem has been solved would be contained 
inside the ANN weights. One of the major issues that ANNs encounter are their lack of understandability. 
A way to better understand solutions given by neural networks would be to know how the inputs are reacting according to the outputs. 
The perfect understanding of the relationship between outputs and inputs is provided by the preimage which gives all corresponding 
inputs for given outputs. The issue that remains is how to compute it. \\\\

Finding inputs for any neural network’s outputs has already been done by giving one approximate and unique solution from the entire set of
corresponding inputs \cite{dua2000inversion}.\\
The most natural method to approximate inverse images of a FFNN output is by retro-propagating it on inputs using gradient descent while
freezing all biases and weights. More recently, researchers built neural networks
in order to invert other neural networks for microwave design filter application
\cite{kabir2008neural}. However, it still doesn’t give the entire set of solutions which leaves the
inverse modeling problem unsolved. Note that these solutions can sometimes
not be used in practice because of constraints on inputs. The problem which
remains is the nonuniqueness of the preimage which has an infinite number
of solutions when the number of inputs is strictly higher than the number of
outputs. It leads every current method to return a unique approximation of a preimage solution for a given output, 
even if the proposed solution is not compatible
with the problem constraints. As a consequence, because many ANNs have an infinity of corresponding inputs for their 
outputs, it is impossible to approximate the preimage numerically.\\
The goal that remains is to find the preimage of the ANN which would give the
entire set of solutions for any outputs. In order to compute a complete preimage, it has to be done analytically . We found two
articles which explored this subject. The first focuses on finding the preimage
of a single point \cite{carlsson2016preimage} while the second extends it to compute the preimage of polytopes
\cite{matoba2020computing}.\\\\

This study aims at giving a method to analytically find the exact preimage of any given
FFNN where activation of hidden layers are linear or piecewise linear functions
like ReLU or Leaky ReLU function. Note that every problem that can be modeled by a continuous function can be 
approximated with an ANN using piecewise linear function like ReLU or Parametric ReLU as activation for it hidden layers \cite{hanin2017approximating}.
The point is to set up the basis of ANN preimage computing by giving the details of a fully working and simple method.
The first part of the study focuses on how to analytically solve this preimage problem in a linear single layer case. 
The second shows how to iterate it to give a solution for any linear multilayers network. The third part focuses
on extending to piecewise linear cases and splitting them to linear problems.
The last part is a description of the implied space and time complexity.\\
It will be shown that because of its exponential time complexity, the exact analytical case can’t be used 
for large neural networks in practice. However, the sets of solutions can be voluntarily reduced in order to compute it 
faster and make the algorithm usable in a practical case.
\newpage
\section{Linear cases}

\subsection{Determined system}
Let’s consider a neural network with only two layers : input and output layer with the same number of neurons for both of them.\\
\\
Let $ N \in \mathbb{N^{*}}$, the number of neurons.\\
Let $\textbf{W}\in \mathbb{R}^{N \times N}$, the matrix of weights.\\
Let $\textbf{x},\textbf{y},\textbf{b} \in \mathbb{R}^{N}$ , respectively inputs, outputs and biases.\\
Let $\alpha,\beta \in \mathbb{R}$\\
Let $f:x \mapsto \alpha x + \beta $, the activation of the layer.\\
An output can be expressed as follows : 

\[y_{k} = f(\sum_{i=1}^{N} w_{k,i}x_{i} + b_{k}) ,  k \in [\![1,N]\!] \]

In the case where the activation function is linear, the previous equation can simply be rewritten as follows :

\begin{align}
    y_{k} = (\sum_{i=1}^{N} \alpha w_{k,i}x_{i} + \alpha b_{k} + \frac{\beta}{N} ),  k \in [\![1,N]\!] 
\end{align}

$\forall k,i \in [\![1,N]\!],w_{k,i}'=\alpha w_{k,i},b_{k}'=\alpha b_{k} + \frac{\beta}{N} $

\begin{align*}
    y_{k} = (\sum_{i=1}^{N}  w_{k,i}'x_{i} +  b_{k}' ),  k \in [\![1,N]\!] 
\end{align*}

It means that finding an inverse image of a layer can be reduced as a matrix problem : 

\begin{equation}
    \textbf{W'}\textbf{x}=\textbf{y}-\textbf{b'},\textbf{W'}= 
    \begin{bmatrix}
    w_{1,1}' & w_{1,2}' & \cdots & w_{1,N}' \\
    w_{2,1}' & w_{2,2}' & \cdots & w_{2,N}' \\
    \vdots  & \vdots  & \ddots & \vdots  \\
    w_{N,1}' & w_{N,2}' & \cdots & w_{N,N}' 
    \end{bmatrix}
    \end{equation}

    This can be solved with Gauss elimination in $O(N^3)$ time complexity for the following system :

    \begin{equation*}
        \textbf{S}= \left[\begin{array}{cccc|c}  
        w_{1,1}'& w_{1,2}' & \cdots & w_{1,N}' & y_{1}-b_{1}' \\
        w_{2,1}' & w_{2,2}' & \cdots & w_{2,N}' & y_{2}-b_{2}'\\
        \vdots  & \vdots  & \ddots & \vdots & \vdots\\
        w_{N,1}' & w_{N,2}' & \cdots & w_{N,N}' & y_{N}-b_{N}'
       \end{array}\right]
    \end{equation*}

    The solution for $\textbf{x}$ is a linear combination of values of $\textbf{y}$ plus a constant.\\
    Solving this system will lead to build a function $p$ of dimension $N$ which gives a solution for any output with no need to recall the algorithm. This function
    $p$ will be expressed as:
    \begin{equation*}
        p(y_{1},...,y_{N}) = 
        \begin{bmatrix}
            x_{1}\\
            \vdots \\
            x_{N}\\
        \end{bmatrix}
        =
        \begin{bmatrix}
            (\sum_{i=1}^{N} a_{i,1}y_{i}) + c_{1}\\
            \vdots \\
            (\sum_{i=1}^{N} a_{i,N}y_{i}) + c_{N}\\
        \end{bmatrix}
    \end{equation*}
    \\\
    Where the $a_{i,j}$ and $c_{j}$ are real values and $P$ the exact preimage of $\textbf{S}$.\\
    $p$ can also be written as a matrix product:

    \begin{equation*}
        p(\textbf{y}) = \textbf{Ay} + \textbf{c}
    \end{equation*}
Where $\textbf{A} \in  \mathbb{R}^{N \times N}$ and $\textbf{c} \in \mathbb{R}^{N}.$

\subsection{Overdetermined systems}

Now let’s consider a neural network once again with only inputs and outputs layers, 
where the input layer has $N$ neurons and output layer $M$ neurons where $N$ is strictly higher than $M$.
As previously seen, if the activation is linear the layer can be inverted by writing the problem as if the activation was the identity function.
In order to simplify the notations, the next calculations will be using the identity function.
\\\\
Now $(1)$ and $(2)$ can be rewritten as follows : 
\\
\\
Let $ N,M \in \mathbb{N^{*}},N>M$

\begin{align}
    y_{k} = \sum_{i=1}^{N} w_{k,i}x_{i} + b_{k} ,  k \in [\![1,M]\!]
\end{align}

\begin{equation}
    \textbf{W}\textbf{x}=\textbf{y}-\textbf{b},\textbf{W}= 
    \begin{bmatrix}
    w_{1,1} & w_{1,2} & \cdots & w_{1,N} \\
    w_{2,1} & w_{2,2} & \cdots & w_{2,N} \\
    \vdots  & \vdots  & \ddots & \vdots  \\
    w_{M,1} & w_{M,2} & \cdots & w_{M,N} 
    \end{bmatrix}
    \end{equation}
\\
    Even with more unknown inputs than outputs , 
    it is still possible to find a solution, writing it as a set of dimension $N-M$, considering $N-M$ inputs as free variables for $M$ dependent variables.
    \\
    \\
    As previously, the system can be written a follows : 

    \begin{equation*}
        \textbf{S}= \left[\begin{array}{cccc|c}  
        w_{1,1} & w_{1,2} & \cdots & w_{1,N} & y_{1}-b_{1} \\
        w_{2,1} & w_{2,2} & \cdots & w_{2,N} & y_{2}-b_{2}\\
        \vdots  & \vdots  & \ddots & \vdots & \vdots\\
        w_{M,1} & w_{M,2} & \cdots & w_{M,N} & y_{M}-b_{M}
       \end{array}\right]
    \end{equation*}

    In order to solve this system by Gaussian elimination, $N-M$ free variables must be moved to the right member as $\textbf{S}$ is overdetermined.
    \\\\
    \\
    Let $\textbf{W}_{1} = \begin{bmatrix}
        w_{1,1} & w_{1,2} & \cdots & w_{1,M} \\
        w_{2,1} & w_{2,2} & \cdots & w_{2,M} \\
        \vdots  & \vdots  & \ddots & \vdots  \\
        w_{M,1} & w_{M,2} & \cdots & w_{M,M} 
        \end{bmatrix}$
    ,$\textbf{W}_{2} = \begin{bmatrix}
        w_{1,M+1} & w_{1,M+2} & \cdots & w_{1,N} \\
        w_{2,M+1} & w_{2,M+2} & \cdots & w_{2,N} \\
        \vdots  & \vdots  & \ddots & \vdots  \\
        w_{M,M+1} & w_{M,M+2} & \cdots & w_{M,N} 
        \end{bmatrix}$
    \\\\\\
    Let $\textbf{x}_{1}=\begin{bmatrix}
        x_{1} \\
        \vdots \\
        x_{M} \\
    \end{bmatrix}
    $
    , $\textbf{x}_{2}=\begin{bmatrix}
        x_{M+1} \\
        \vdots \\
        x_{N} \\
    \end{bmatrix}
    $

    \begin{equation*}
        \begin{split}
            \textbf{W}\textbf{x}  =\textbf{y}-\textbf{b} &\Rightarrow 
            \begin{bmatrix}\textbf{W}_{1}&\textbf{W}_{2} \end{bmatrix} \begin{bmatrix}\textbf{x}_{1} \\\textbf{x}_{2} \end{bmatrix} =\textbf{y}-\textbf{b}\\
             & \Rightarrow  \textbf{W}_{1} \textbf{x}_{1} + \textbf{W}_{2} \textbf{x}_{2} = \textbf{y}-\textbf{b} \\
             & \Rightarrow  \textbf{W}_{1} \textbf{x}_{1} = \textbf{y}-\textbf{b}-\textbf{W}_{2} \textbf{x}_{2}
            \end{split}
    \end{equation*}

    The previous system can be rewritten as : 

    \begin{equation*}
        \textbf{S'}= \left[\begin{array}{cccc|c}  
        w_{1,1} & w_{1,2} & \cdots & w_{1,M} & y_{1}-b_{1} - \sum_{i=M+1}^{N} w_{1,i}x_{i}\\
        w_{2,1} & w_{2,2} & \cdots & w_{2,M} & y_{2}-b_{2} - \sum_{i=M+1}^{N} w_{2,i}x_{i}\\
        \vdots  & \vdots  & \ddots & \vdots & \vdots\\
        w_{M,1} & w_{M,2} & \cdots & w_{M,M} & y_{M}-b_{M} - \sum_{i=M+1}^{N} w_{M,i}x_{i}
       \end{array}\right]
    \end{equation*}
    Under this form the values of $\textbf{x}_{2}$ can now be considered as new free variables of the system such as those of $\textbf{y}$.\\
    As $\textbf{S'}$ is now a determined system it can be solved by Gaussian elimination.
    The preimage will be a function $p$ of dimension $N$ that depends on the $y_{i \in [\![1,M]\!]}$ 
    and also on the new real free variables of $x_{2}$\\
    \begin{equation*}
    \begin{split}
        p(y_{1},...,y_{M}) &= 
        \begin{bmatrix}
            (\sum_{i=1}^{M} a_{1,i,1}y_{i})+ (\sum_{i=M+1}^{N} a_{1,i,2}x_{i})  + (\sum_{i=1}^{N} a_{1,i,3}b_{i})\\
            \vdots \\
            (\sum_{i=1}^{M} a_{N,i,1}y_{i})+ (\sum_{i=M+1}^{N} a_{N,i,2}x_{i})  + (\sum_{i=1}^{N} a_{N,i,3}b_{i}\\
        \end{bmatrix}\\
        \Rightarrow p(\textbf{y})&= \textbf{A}_{1}\textbf{y} + \textbf{A}_{2}\textbf{x}_{2}+ \textbf{A}_{3}\textbf{b}
    \end{split}
    \end{equation*}
    Where $\textbf{A}_{1} \in  \mathbb{R}^{N \times M}$,$\textbf{A}_{2} \in  \mathbb{R}^{N \times (N-M)}$ and $\textbf{A}_{3} \in \mathbb{R}^{N}.$

\subsection{Underdetermined systems}

In the case where $M$ is strictly higher than $N$ , the system is underdetermined.\\
Let $\mathbf{X}$ the space of inputs and $\mathbf{Y}$ the space of outputs.

\begin{equation*}
    \begin{split}
    f(X) \subset Y & \Rightarrow dim f(X) + dim \{x, f(x) = b\}  = dim X \\
    & \Rightarrow dim f(X) \leq dim X = N <  M
    \end{split}
\end{equation*}

If the system has more free variables than inputs and no hidden layer 
with strictly less neurons than the output layer then the system can be solved as follows :
\\
\\
Let $ N,M \in \mathbb{N^{*}},N<M$\\
Let $F \geq M$, the total number of free variables.
\\ Let $\textbf{A} \in \mathbb{R}^{M \times F}$

\begin{equation}
    \textbf{Wx}=\textbf{Ay}-\textbf{b}
\end{equation}
\\
Let $\textbf{A}_{1} \in \mathbb{R}^{M \times (M-N)},\textbf{A}_{2} \in \mathbb{R}^{M \times (F+N-M)}, 
\textbf{A} = \begin{bmatrix}\textbf{A}_{1}&\textbf{A}_{2} \end{bmatrix}$\\
Let $\textbf{y}_{1} \in \mathbb{R}^{M-N},\textbf{y}_{2} \in \mathbb{R}^{N},
\textbf{y} = \begin{bmatrix}\textbf{y}_{1} \\ \textbf{y}_{2} \end{bmatrix}$

\begin{equation}
    \begin{split}
        \textbf{W}\textbf{x}  =\textbf{Ay}-\textbf{b} &\Rightarrow 
        \textbf{W}\textbf{x}  = \textbf{A}_{1}\textbf{y}_{1}  + \textbf{A}_{2}\textbf{y}_{2} -\textbf{b}\\
         & \Rightarrow  \textbf{W}\textbf{x}-\textbf{A}_{1}\textbf{y}_{1} = \textbf{A}_{2}\textbf{y}_{2}-\textbf{b} \\
         & \Rightarrow \begin{bmatrix}\textbf{W}&\textbf{A}_{1} \end{bmatrix} \begin{bmatrix}\textbf{x} \\\textbf{y}_{1} 
        \end{bmatrix} =\textbf{A}_{2}\textbf{y}_{2}-\textbf{b}
        \end{split}
\end{equation} 
\\
\
Let $\textbf{W}'= \begin{bmatrix}\textbf{W}&\textbf{A}_{1} \end{bmatrix},\textbf{x}' = \begin{bmatrix}\textbf{x} \\\textbf{y}_{2} \end{bmatrix}$
\\\\
\begin{equation}
    (6) \Rightarrow \textbf{W}'\textbf{x}' =\textbf{A}_{2}\textbf{y}_{2}-\textbf{b}
\end{equation}
\\
\\
The system $(5)$ can now be rewritten by adding $M-N$ variables of $\textbf{y}$ as unknown variables of $(7)$.\\ 
Once again, the system $(7)$ can be solved thanks to Gaussian elimination with $M$ solutions of $D=F+N-M$ dimensions.
The preimage $p$ will have the same form as the determined case.

\begin{equation*}
p(\textbf{y}_{2})= \textbf{A}_{3}\textbf{y}_{2} + \textbf{A}_{4}\textbf{b}
\end{equation*}

Where $\textbf{A}_{3} \in  \mathbb{R}^{M \times (F+N-M)}$ and $\textbf{A}_{4} \in  \mathbb{R}^{M}.$

\subsection{Underdetermined system without extra free variables}

A specific case occurs when the dimension of implied output space is strictly higher than input space. 
In fact, when the output layer has strictly more neurons than the input layer, most of the target outputs would be not valid 
and will have no solution in the preimage set. However, it is still possible to find the closest valid output in terms of least squares error
in order to perform the previously shown algorithm to find a corresponding input. The following section will quickly shows how to applied
least squared error to solve underdetermined systems. \\\\
Let $N,M \in \mathbb{N^{*}},N<M$, respectively the inputs and outputs length.\\
Let $\textbf{W}\in \mathbb{R}^{M \times N}$.\\
Let $\textbf{y} \in \mathbb{R}^{M}$, the outputs vector.\\
Let $\textbf{x} \in \mathbb{R}^{N}$, the inputs vector.

\begin{equation*}
    \textbf{Wx}= \textbf{y}
\end{equation*}

This system can't be solved for $\textbf{x}$ in a general case. However it is still possible to find the vector $\widehat{\textbf{y}}$ that minimizes
the least squares criterion for $\textbf{y}$. Then the system can be solved by simple Gaussian Elimination and will returns the closest solution in terms
of least squares error.\\\\
Let $\boldsymbol{\varepsilon} \in \mathbb{R}^{M}, \widehat{\textbf{y}} = \textbf{y}+ \boldsymbol{\varepsilon}$, the error vector.\\
The goal is now the find the smallest corresponding $\boldsymbol{\varepsilon}$ in terms of $L2$-norm.\\
Let rewrite the system such as :

\begin{equation*}
    \textbf{Wx}= \widehat{\textbf{y}} = \textbf{y} + \boldsymbol{\varepsilon}
\end{equation*}
Because $M$ is strictly higher than $N$, there are $M-N$ missing variables to keep the system determined.
These $M-N$ variables can be taken from $\boldsymbol{\varepsilon}$ and treated as unknown.
\\Let $ \boldsymbol{\varepsilon}_{1} \in \mathbb{R}^{M-N}, \boldsymbol{\varepsilon}_{2} \in \mathbb{R}^{N},\boldsymbol{\varepsilon} =$
$
\begin{bmatrix}
    \boldsymbol{\varepsilon}_{1}
    & \boldsymbol{\varepsilon}_{2}
\end{bmatrix}^{T}
$
\begin{equation*}
    \textbf{Wx}
    = 
    \textbf{y}
     + 
     \begin{bmatrix}
        \boldsymbol{\varepsilon}_{1}
        & \boldsymbol{\varepsilon}_{2}
    \end{bmatrix}^{T} 
    \Rightarrow
    \begin{bmatrix}
        \textbf{W}
        & 
        \begin{matrix}
        -\textbf{I}_{M-N}\\
        \textbf{0}_{N,M-N}
        \end{matrix}
    \end{bmatrix}
    \textbf{x} =
    \textbf{y}
     + 
     \begin{bmatrix}
        \textbf{0}_{1,N}
        & \boldsymbol{\varepsilon}_{2}
    \end{bmatrix}^{T} 
\end{equation*}
Under this form the system becomes determined and can be solved by Gaussian Elimination.
The two vector $\textbf{x}$ and $\boldsymbol{\varepsilon}_{1}$ will be expressed as linear combination of 
$\textbf{y}$ and $\boldsymbol{\varepsilon}_{2}$. \\
Because, $\widehat{\textbf{y}}$ is the closest valid output of $\textbf{y}$ in terms of least squares,
the $L2$-norm of $\boldsymbol{\varepsilon}$ have to be minimized.\\\\
Let $\textbf{A} \in \mathbb{R}^{(M-N) \times M},\textbf{B} \in \mathbb{R}^{(M-N) \times N}, 
\boldsymbol{\varepsilon}_{1} = \textbf{A}\textbf{y} + \textbf{B}\boldsymbol{\varepsilon}_{2}$.\\
\begin{equation*}
    \begin{split}
    \boldsymbol{\varepsilon}^{T}\boldsymbol{\varepsilon} &=
    \begin{bmatrix}
        \boldsymbol{\varepsilon}_{1}^{T}
        & \boldsymbol{\varepsilon}_{2}^{T}
    \end{bmatrix}
    \begin{bmatrix}
        \boldsymbol{\varepsilon}_{1}
        \\ \boldsymbol{\varepsilon}_{2}
    \end{bmatrix}\\
    &=\boldsymbol{\varepsilon}_{1}^{T}\boldsymbol{\varepsilon}_{1} + \boldsymbol{\varepsilon}_{2}^{T}\boldsymbol{\varepsilon}_{2}
    + 2\boldsymbol{\varepsilon}_{2}^{T}\boldsymbol{\varepsilon}_{1}\\
    &= 
    \begin{bmatrix}
    \textbf{A}\textbf{y} + \textbf{B}\boldsymbol{\varepsilon}_{2}
    \end{bmatrix}^{T}
    \begin{bmatrix}
        \textbf{A}\textbf{y} + \textbf{B}\boldsymbol{\varepsilon}_{2}
    \end{bmatrix} 
    + \boldsymbol{\varepsilon}_{2}^{T}\boldsymbol{\varepsilon}_{2}
    +2\boldsymbol{\varepsilon}_{2}^{T}
    \begin{bmatrix}
        \textbf{A}\textbf{y} + \textbf{B}\boldsymbol{\varepsilon}_{2}
        \end{bmatrix}\\
    &= 
    \begin{bmatrix}
        \textbf{A}\textbf{y} + (\textbf{B}+\textbf{I}_{n})\boldsymbol{\varepsilon}_{2}
    \end{bmatrix}^{T}
    \begin{bmatrix}
        \textbf{A}\textbf{y} + (\textbf{B}+\textbf{I}_{n})\boldsymbol{\varepsilon}_{2}
    \end{bmatrix}
\end{split}
\end{equation*}
The next step is to find the corresponding $\boldsymbol{\varepsilon}_{2}$ that minimize the previous expression.\\
Let $f:\mathbb{R}^{N} \longrightarrow \mathbb{R},f(\textbf{x})=
\begin{bmatrix}
    \textbf{A}\textbf{y} + (\textbf{B}+\textbf{I}_{n})\textbf{x}
\end{bmatrix}^{T}
\begin{bmatrix}
    \textbf{A}\textbf{y} + (\textbf{B}+\textbf{I}_{n})\textbf{x}
\end{bmatrix}
$
\begin{equation*}
    \begin{split}
    &\frac{\partial f}{\partial \textbf{x}}(\textbf{x})=2
    \begin{bmatrix}
        \textbf{A}\textbf{y} + (\textbf{B}+\textbf{I}_{n})\textbf{x}
    \end{bmatrix}^{T}
    \begin{bmatrix}
        \textbf{B}+\textbf{I}_{n}
    \end{bmatrix}\\
    &\frac{\partial^{2} f}{\partial \textbf{x}^{2}}(\textbf{x})=2
    \begin{bmatrix}
        \textbf{B}+\textbf{I}_{n}
    \end{bmatrix}^{T}
    \begin{bmatrix}
        \textbf{B}+\textbf{I}_{n}
    \end{bmatrix}>0
    \end{split}
\end{equation*}
Because the second derivative of $f$ is positive, finding the $\textbf{x}$ that minimizes it is equivalent to find thezeros of its derivative.
\begin{equation*}
    \begin{split}
    \frac{\partial f}{\partial \textbf{x}}(\textbf{x})=0 &\Rightarrow
    2
    \begin{bmatrix}
        \textbf{A}\textbf{y} + (\textbf{B}+\textbf{I}_{n})\textbf{x}
    \end{bmatrix}^{T}
    \begin{bmatrix}
        \textbf{B}+\textbf{I}_{n}
    \end{bmatrix}=0\\
    &\Rightarrow
    (\textbf{B}+\textbf{I}_{n})\textbf{x} =-\textbf{A}\textbf{y} 
    \end{split}
\end{equation*}
The equation can be solved under this form thanks to Gaussian Elimination.

\subsection{Multilayers network}
Inverting a multilayer network only needs to iterate the previous method on each layer to find the exact preimage. 
The inputs at each iteration will be the previous outputs. \\This comes from the fact that for multivariate functions of $\mathbb{R}$, 
the preimage of the composition is the composition of the preimages \cite{matoba2020computing}.
\newpage
\section{Piecewise linear cases}

The previous section shows how to compute the exact preimage of a multilayer FFNN with linear activation. 
The same problem can be solved in case of nonlinear activation under particular constraints.
\\
\\
Let $f$ a nonlinear activation function which is bijective in at least one interval $I$.
\\
Let $f^{-1}$ the inverse function of $f$ in $I$.
\\
Let $ N,M \in \mathbb{N^{*}}$.
\\
Considering the same notations as previously :
\
\begin{equation*}
    y_{k} = f(\sum_{i=1}^{N} w_{k,i}x_{i} + b_{k}) ,  k \in [\![1,M]\!] 
    \Rightarrow f^{-1}(y_{k}) = \sum_{i=1}^{N} w_{k,i}x_{i} + b_{k} ,  k \in [\![1,M]\!]
\end{equation*}
\\
The system $\textbf{S}$ describes the inverse images set of any layer in a FFNN where there is more neurons in the input layer than in the output layer :
\\\\
\begin{equation}
    \textbf{S}= \left[\begin{array}{cccc|c}  
    w_{1,1} & w_{1,2} & \cdots & w_{1,N} & f^{-1}(y_{1)}-b_{1} \\
    w_{2,1} & w_{2,2} & \cdots & w_{2,N} & f^{-1}(y_{2})-b_{2}\\
    \vdots  & \vdots  & \ddots & \vdots & \vdots\\
    w_{M,1} & w_{M,2} & \cdots & w_{M,N} & f^{-1}(y_{M})-b_{M}
   \end{array}\right]
\end{equation}
\\
\\
This system cannot be solved in the general case where $f^{-1}$ is nonlinear and $dim(y_{i})\geqslant 1,i \in [\![1,M]\!]$.
However, when $y_{i},f^{-1}(y_{i}) \in \mathbb{R} $ , the system $(8)$ can be solved by simple Gaussian elimination.
It means that any function $f$ where $f^{-1}:\mathbb{R} \longrightarrow \mathbb{R}$ exists can be used as the output activation of 
a network without preventing it from being inverted.

\subsection{Piecewise linear activation : Parametric ReLU }
Specific nonlinear cases can be solved if the activation function is piecewise linear.
\\
\\
As a reminder :

\begin{equation*}
    PReLU : \mathbb{R} \longrightarrow \mathbb{R}
\end{equation*}

\[
    PReLU(x)= 
\begin{cases}
    x,& \text{if } x>0\\
    \alpha x, & \text{otherwise}
\end{cases}
\]
\\
\\
Let's now consider the inverse function of PReLU:
\\
\\
Let $\alpha \geqslant 0$
\\
Let $iPRL : \mathbb{R} \longrightarrow \mathbb{R}$, the inverse function of Parametric ReLU of parameter $\alpha$.

\[
    iPRL(y)= 
\begin{cases}
    y,& \text{if } y>0\\
    \frac{y}{\alpha}, & \text{otherwise}
\end{cases}
\]

The only thing which changes from the linear case is that the truth value of $y > 0$ must be known.
The answer is trivial if $y$ is known. If $y$ is expressed as a linear combination of free variables from previous equations,
this problem can be reduced as a linear inequalities system :

\begin{equation}
\begin{cases}
    (\sum_{i=1}^{N} w_{i,1}x_{i}) + b_{1} \geqslant  0
    \\\\ (\sum_{i=1}^{N} w_{i,2}x_{i}) + b_{2} \geqslant 0
    \\ \vdots 
    \\ (\sum_{i=1}^{N} w_{i,M}x_{i}) + b_{M} \geqslant  0
\end{cases}
\Rightarrow
\textbf{WX} + \textbf{b} \geqslant  0
\end{equation}
\\\\
Notice that there are $2^{M}$ different systems to solve depending on whether $y_{j}>0$ or $y_{j}<0$.
In this study, these systems were all solved independently.
Also notice that these systems can all be written with left member higher than $0$ by multiplying it by $-1$ 
which will reverse the sign of the inequation as an effect.
\\\\
\
This previous section shows that the method will work with ReLU activation if any system of multivariate linear inequalities can be analytically solved.
The next section explains how to solve such systems.

\subsection{Solving systems of multivariates linear inequalities }
The goal of this section is to analytically solve multivariate linear inequalities systems.
\paragraph{Expression of the problem} \
\\\\
Let $N,M \in \mathbb{N}^{*}$\\
Let $\{w_{i,j}\}_{i \in [\![1,N]\!],j \in [\![1,M]\!]} \in \mathbb{R}^{N+M}$, a family of real constants.\\
Let $\{b_{j}\}_{j \in [\![1,M]\!]} \in \mathbb{R}^{M}$, another family of real constants.\\
Let $k \in [\![1,M]\!]$
\\\\
In a general case, inequalities can be written with the following form :\\

\begin{equation}
(\sum_{i=1}^{N} w_{i,k}x_{i}) + b_{k} \geqslant 0
\end{equation}
\\
Considering $(10)$, a system of multivariates linear inequalities can be described as follows in the general case:
\\
\begin{equation}
    S=
    \begin{cases}
         (\sum_{i=1}^{N} w_{i,1}x_{i}) + b_{1} \geqslant 0
        \\  (\sum_{i=1}^{N} w_{i,2}x_{i}) + b_{2} \geqslant 0
        \\ \vdots 
        \\  (\sum_{i=1}^{N} w_{i,M}x_{i}) + b_{M} \geqslant 0
    \end{cases}
    \end{equation}
\\
    Note that large inequalities can be replaced by strict inequalities without changing the method.

\paragraph{Analytical Algorithms for solving linear inequalities systems} \ \\

In general case, linear inequalities system can be written as: 
\begin{equation*}
\textbf{W}\textbf{x}\geqslant0 
\end{equation*}

A classic algorithm to analytically solve this kind of system is Fourier-Motzkin Elimination \cite{motzkin1952theory}.
This algorithm is simple but has the downside of being extremely slow in terms of time complexity because of adding multiple redundant 
inequalities. However, from a theoretical point of view, such an algorithm still provides a constructivist way to analytically solve linear inequalities
systems. \\ The details of Fourier-Motzkin Elimination algorithm is given in Appendix A.

\paragraph{Solutions} \ \\  

The analytical solutions of a multivariate linear inequalities system solve by Fourier-Motzkin Elimination or any other analytical algorithm will 
looks as follows :\\\\
Let $N,M \in \mathbb{N}^{*}$\\
Let $\textbf{W} \in \mathbb{R}^{N \times M}$ a squared matrix of real constants.\\
Let $\textbf{x} \in \mathbb{R}^{M}$ a vector of real variables.\\
Let $\textbf{S}$ the solution of the following system of multivariate linear inequalities : 
\begin{equation*}
    \textbf{W}\textbf{x}\geqslant0
\end{equation*}
Let $j_{1},j_{2} \in [\![0,M]\!], M = j_{1} +j_{2} $, respectively the number of constrained and unconstrained variables.\\
Let $\{L_{i}\}_{i \in [\![1,j_{1}]\!]},\{G_{i}\}_{i \in [\![1,j_{1}]\!]}$ respectively 
the sets of constraints that have to be lower and greater than a given $x_{i}$.

\begin{equation*}
    \begin{split}
    \textbf{S}=\{x_{i}\}_{i \in [\![1,M]\!]},&
    \begin{cases}
        min(L_{1}) \leqslant x_{1} \leqslant max(G_{1})\\
        \vdots   \\
        min(L_{j_{1}}) \leqslant x_{j_{1}} \leqslant max(G_{j_{1}})\\
        -\infty < x_{j_{1}+1} < +\infty\\
        \vdots   \\
        -\infty < x_{j_{1}+j_{2}} < +\infty\\
    \end{cases}\\
    = \textbf{x} ,&
    \begin{cases}
        min(\textbf{L}) \leqslant \textbf{x}_{1} \leqslant max(\textbf{G})\\
        -\infty < \textbf{x}_{2} < +\infty
    \end{cases}
    \end{split}
\end{equation*}
Let $i \in [\![1,j_{1}]\!]$ such as $L_{i},G_{i}$ are the $i$-th component of $\textbf{L},\textbf{G}$.\\
Let $n \in \mathbb{N}$, number of constraints for $L_{i}$.\\
Let $\textbf{A} \in \mathbb{R}^{ n \times M }$, the coefficient matrix of the constraints variables in $L_{i}$.\\
Let $\textbf{B} \in \mathbb{R}^{ n \times M }$, the coefficient matrix of the unconstraints variables in $L_{i}$.\\
$\forall h \in [\![1,n]\!], \forall k \in [\![1,M]\!], k \geqslant j_{1}, a_{h,k} =0 $.\\
Respectively, $\forall h \in [\![1,n]\!], \forall k \in [\![1,M]\!], k > j_{2}, b_{h,k} =0 $.\\
Let $\textbf{c} \in \mathbb{R}^{n}$, a vector of real constants.\\
The expression of $G_{i}$ will be symmetrical to $L_{i}$.

\begin{equation*}
    \begin{split}
    L_{i}&=
    \begin{bmatrix}
        \sum_{k=1}^{i-1} a_{1,k}x_{k} + \sum_{k=1}^{j_{2}}b_{1,k}x_{k} + c_{1} 
        & \cdots 
        & \sum_{k=1}^{i-1} a_{n,k}x_{k} + \sum_{k=1}^{j_{2}}b_{n,k}x_{k} + c_{n}
    \end{bmatrix}^{T}\\
    &=
    (\textbf{A} + \textbf{B})\textbf{x} + \textbf{c}
\end{split}
\end{equation*}

As seen previously, it is possible to solve any systems of multivariate linear inequalities analytically.
It means that the last step to reverse the layer is to compute each solution by solving each $2^{M}$ different system. 
After that, the next layer can be inverse thanks to the same method. Notice that this time, variables would be under constraints, 
but it doesn't change the method as these constraints can be added as new inequalities in the next system to solve.

\subsection{Application to piecewise linear case}
Let $ N,M \in \mathbb{N^{*}}$, the number of inputs and outputs neurons.\\
Let $\textbf{W}\in \mathbb{R}^{N \times M}$, the matrix of weights.\\
Let $\textbf{x},\textbf{b} \in \mathbb{R}^{N}$ , respectively inputs and bias.\\
Let $\textbf{y} \in \mathbb{R}^{M}$, the outputs where $\textbf{y}$ is a vector of $N$ variables (free or under constraints).\\\\
The preimage of the network can be computed by iteratively solve the following equation for each layer :

\begin{equation*}
\textbf{W}\textbf{x}=\textbf{y}-\textbf{b}
\end{equation*}
Note that because the $y_{i \in [\![1,N]\!]}$ are variables, they have to be kept as symbols during the entire computation. Thanks to the previous result, 
multivariate inequalities systems can be solved in a general case. It means that for each layer of size $k \in \mathbb{N}^{*}$, the $2^{k}$ different inequalities systems 
can be analytically solved. Solving these systems will lead to create $2^{k}$ new sets of solutions that will have to be taken as new starting points for the next iteration
of layer preimage finding. 
\\\\ \
\textbf{Example: }\\
Let's consider a neural network with piecewise linear activation for each hidden layers.\\
Let $N,M \in \mathbb{N^{*}},M \geqslant N$, respectively the length of the inputs and outputs layer.\\
Let $H \in \mathbb{N}$, the number of hidden layers. \\
Let $\{N_{k}\}_{k \in [\![1,H]\!]}\in \mathbb{N}^{H}$, the family that contains the length of each $k$-th hidden layers.\\\\
As previously seen,the $k$-th hidden layer introduces $2^{N_{k}}$ inequalities systems to solve which lead to create $2^{N_{k}}$ new sets of solutions with their constraints. 
By direct induction, the total number of solution sets for the neural network $\Omega$ will be expressed as :  
\begin{equation*}
\Omega=2^{N+\sum_{k=1}^{H} N_{k}}
\end{equation*}
Knowing this, the last step is to express the form of the preimage.\\
The $P_{k \in [\![1,\Omega]\!]}$ solutions sets of $\textbf{P}$  can be written as follows:\\\\
Let $\textbf{W}\in \mathbb{R}^{M \times N \times \Omega}$, the matrix of coefficients for the $y$.\\
Let $\textbf{c} \in \mathbb{R}^{N \times \Omega}$ , the constants coefficients.\\
Let $\textbf{y} \in \mathbb{R}^{M}$, the outputs vector\\
The preimage $\textbf{P}$ will depend of the values of the $y_{i \in [\![1,N]\!]}$ and will have constraints given by the $\tau_{j \in [\![1,M-N]\!]}$.\\
\\
Let $T' \in \mathbb{N}$, the number of variables $\tau$ that were eliminated during the whole computation of the preimage. \\
Let $T = M-N-T'$, the number of $\tau$ that still has free variables.\\
Let $\textbf{A} \in \mathbb{R}^{T \times N \times \Omega}$, the coefficients matrix for the $\tau$.\\
Let $\boldsymbol{\tau} \in \mathbb{R}^{T \times \Omega}$, the extra free variable.
Let $\boldsymbol{\tau}' \in \mathbb{R}^{T' \times \Omega}$, the vector of the eliminated $\tau$ that became constants real constraints of the solutions.\\
Let $\textbf{L},\textbf{G},\textbf{L'},\textbf{G'}$ respectively the sets of constraints that have to be lower and greater than a given $\tau$ and $\tau'$.\\
\\
Let  $k \in [\![1,\Omega]\!]$, the index of the solution.\\
$\textbf{P}$ is a set of dimension $\Omega$ per $N$.\\
The $k$-th solution of $\textbf{P}$  will be expressed as : 

\begin{equation*}
P_{k}(y_{1},...,y_{M})=
\begin{bmatrix}
(\sum_{i=1}^{M} w_{i,1,k}y_{i} )+ (\sum_{i=1}^{T} a_{i,1,k}\tau_{i,k}) + c_{1,k} \\
\vdots \\
(\sum_{i=1}^{M} w_{i,N,k}y_{i} )+ (\sum_{i=1}^{T} a_{i,N,k}\tau_{i,k}) + c_{N,k}  \\
\end{bmatrix}
,
\begin{cases}
    min(L_{1,k})  \leqslant \tau_{1,k} \leqslant max(G_{1,k})\\
    \vdots   \\
    min(L_{T,k})  \leqslant \tau_{T,k} \leqslant max(G_{T,k})\\
    min(L'_{1,k})  \leqslant\tau'_{1,k} \leqslant max(G'_{1,k})\\
    \vdots   \\
    min(L'_{T',k})  \leqslant \tau'_{T',k} \leqslant max(G'_{T',k})
\end{cases}
\end{equation*}

Which can be simplify as :
\begin{equation*}
    P_{k}(\textbf{y})= \textbf{W}_{k}\textbf{y} + \textbf{A}_{k}\boldsymbol{\tau}_{k} + \textbf{c}_{k},
    \begin{cases}
        \textbf{L}_{k}  \leqslant \boldsymbol{\tau}_{k} \leqslant \textbf{G}_{k}\\
        \textbf{L'}_{k}  \leqslant \boldsymbol{\tau'}_{k} \leqslant \textbf{G'}_{k}
    \end{cases}
\end{equation*}

\subsection{Extension to simple ReLU activation}

ReLU preimage can be described as follow :
\\
\\
As a reminder :

\begin{equation*}
    ReLU : \mathbb{R} \longrightarrow \mathbb{R^{+}}
\end{equation*}

\[
    ReLU(x)= 
\begin{cases}
    x,& \text{if } x>0\\
    0, & \text{otherwise}
\end{cases}
\]
\\
\\

Let $iRL : \mathbb{R} \longrightarrow \mathbb{R^\mathbb{R}}$, the preimage of ReLU.

\[
    iRL(y)= 
\begin{cases}
    y,& \text{if } y>0\\
            ]-\infty,0],              & \text{otherwise}
\end{cases}
\]
\\
Inverting a ReLU layer is possible thanks to the same method as previously seen.
In addition to it, a negative real variable must be added for each inequality solved as less than zero for a given set.

\newpage
\section{Generalized Preimage}

Neural networks can be described as ordered weights associated with activation function. Each layer's weights can be formalized as 
matrices. It means that all ANN weights can be expressed as a vector $n+1$ matrices, 
where $n$ is the number of hidden layers.\\\\
Let $\{\textbf{W}_{i}\}_{i \in [\![1,n+1]\!]}$, the family of weights matrice of a neural network.\\
Let $\textbf{W}=
\begin{bmatrix}
    \textbf{W}_{1} & \cdots & \textbf{W}_{n+1} 
\end{bmatrix}^{\textbf{T}}$, the vector of weight matrices.\\\\
The generalized preimage $P(\textbf{y},\textbf{W})$ is an expression of the preimage of all neural networks that have the same activation functions
and also the same dimension for the matrix $\textbf{W}$ and it element.\\
Notice that $P$ can be computed using the same way as seen previously. It can be done by considering each weight as a variable instead of a constant.
The only information that has to be known is the sign of each weight. This is mandatory during the elimination
of variables when solving inequalities system.\\\\
$\forall i \in [\![1,n+1]\!], \forall w \in \textbf{W}_{i} , w' = sgn(w)\lvert w\rvert$, where $sgn$ is the sign function. \\\\
If $sgn(w)$ is known then the preimage can be computed by considering $\lvert w\rvert$ as a positive variable.
Note that the exact same function will result of this computation for neural network with same sign for each weight.\\
This result shows that neural networks with the same sign for each weight have similar properties such as the same generalized preimage.

\newpage
\section{Complexity}
\subsection{Space Complexity}\
\\
The space complexity is directly linked to the number of solutions. In a piecewise linear case, the number of solutions has been given in
section $3.3$ which was equal to $\Omega=2^{N+\sum_{k=1}^{H} N_{k}}$,where $N$ is the length of the outputs layer, the $N_{k}$ of the $k$-th hidden layer and
$H$ the total number of hidden layer.
It gives a space complexity of $O(\Omega)=O(2^{N+\sum_{k=1}^{H} N_{k}})$.\\
The algorithm space complexity is exponential.

\subsection{Time Complexity}\
\\
The algorithm time complexity depends on the linear inequalities solver. Let $f$ the time complexity function of the linear inequalities systems solver.\\
Let $n,m$ respectively the number of inequalities and its number of variables. Then $f$ is a function of $n,m$ such that the time complexity is equal to 
$O(f(n,m))$.\\
For a given multilayer neural network, let $S$ the sum of the length of the hidden layers and the output layer.\\
Let $L,N$ respectively be the length of the first hidden layer of the output layer.\\
Then the time complexity is equal to : $ O(2^{S}f(L,L-N))$. In that case, $ 2^{S}$ is the total number of system to solve for the first hidden layer, and
$L-N$ the number of free variables.\\\\

The algorithm depends on the Fourier-Motzkin time complexity. The Fourier-Motzkin time complexity is in $O(n^{2^{m}})$ in the worst case
(Where $n$ is number of inequalities and $m$ the number of variables to eliminate) \cite{motzkin1952theory}. Because of the number of solution that growth 
exponentially, the time complexity of the algorithm is expressed as : $O(2^{S}L^{2^{L-N}})$, where $S$ is the sum of the length of the hidden layer layers
and the input layer, $N$ the length of the output layer, $L$ the length of the first hidden layer.\\
The total time complexity of the algorithm is double exponential.
\newpage
\section{Conclusion}

Computing the preimage of ANN with linear activation for hidden layers can be reduced to a matrix problem. 
It ensures that solution sets can be found thanks to analytical calculation and Gaussian elimination. 
For the nonlinear case, the method can’t invert networks with non-piecewise linear hidden layers activation like tanh or sigmoid in general cases. 
Notice that it is still possible in specific cases or if the non-piecewise linear activation concern the output layer.\\ 
With piecewise linear functions like ReLU or Parametric ReLU, it become possible to analytically describe the preimage of
any kind of FFNN. This has been done thanks to the fact that preimages of piecewise linear functions are piecewise linear themselves and can
be separated into linear subproblems if the sign of their inputs are known. These signs can be found by solving a system of linear 
inequalities thanks to Fourier-Motzkin elimination. Every analytical solver for multivariate linear inequalities system would 
have the exact same result.\\ Notice that even with more outputs than inputs, the method can find and return the preimage of the closest valid output. 
This can be done through least squares error minimization.\\\\
This study shows that ANN that can approximate any continuous function can be fully inverted analytically. 
Due to the exponential time complexity for the computation of the preimage because of the growing number of sets of solutions, 
it can seem that this approach is not usable in practice for large neural networks. However, for practical application, most of the time only a
small part of all solutions will be necessary. It means that computation time may be reduced by not computing the entire set of solutions. \\\\
Exact preimage computation will give a better understanding of choices taken by FFNN. It also enable to better control 
the behavior of ANN that are often used as blackbox function.\\ As a resume , this study shows a method that setup a base for FFNN preimage 
computation thanks to a straight-forward algorithm for analytical solving. It ensures that exact preimages of FFNN can be theoretically 
found and open the way to other ANN inverting based studies.
\\\\\\
Please find the associated code : \url{https://github.com/TheoN70/preimage_nn}

\newpage
\printbibliography
\newpage
\begin{appendices}
\section{Fourier-Motzkin Elimination}
Fourier-Motzkin elimination \cite{motzkin1952theory} is an algorithm which eliminate variables in multivariate linear inequalities systems.
\\
In a general case, linear inequalities system can be written as: 

\begin{equation*}
\textbf{W}\textbf{x}\geqslant0 
\end{equation*}

\begin{algorithm}
    \caption{Fourier-Motzkin algorithm} 
    \hspace*{\algorithmicindent} \textbf{Inputs} : \textbf{W},\textbf{x}
    
    \begin{algorithmic} 
        
    \WHILE{$\exists (w_{i_1,j},w_{i_2,j}),(i_1,i_2) \in [\![1,N]\!]^{2},j \in [\![1,M]\!], w_{i_1,j}w_{i_2,j}<0)$}
    \STATE $plus \gets [];$ (Where $[]$ is an empty list)
    \STATE $minus \gets [];$
    \STATE $other \gets [];$
    \FOR{$i \gets 1 $ to $N$}
    \IF{$w_{i,j} \neq 0$}
    \IF{$w_{i,j} > 0$}
    \STATE $plus$ add $(\frac{1}{w_{i,j}})W_{i};$ (Where $W_{i}$ is the $i$-th lign of the matrix \textbf{W})
    \ELSE 
    \STATE $minus$ add $(\frac{1}{w_{i,j}})W_{i};$
    \ENDIF
    \ELSE
    \STATE $other$ add $W_{i};$
    \ENDIF
    \ENDFOR
    \STATE $N \gets card(minus)card(plus) + card(other);$ ($card$ the cardinal function)
    \STATE $M \gets M-1;$
    \STATE $\textbf{W}\gets 0_{N,M}$
    \STATE $i \gets 1;$
    \FOR{$eq_{1}$ in $plus$}
    \FOR{$eq_{2}$ in $minus$}
    \STATE $W_{i} \gets eq_{1}-eq_{2}$
    \STATE $i \gets i+1;$
    \ENDFOR
    \ENDFOR
    \FOR{$eq$ in $other$}
    \STATE $W_{i} \gets eq$
    \STATE $i \gets i+1;$
    \ENDFOR
    \ENDWHILE 
    \RETURN \textbf{W}
    \end{algorithmic}
    \end{algorithm}
    \
    \\
    Let $k \in [\![1,N]\!]$\\
    Let $P,Q \in \mathbb{N}, P+Q=M$ \\\\
    With Fourier-Motzkin algorithm, each eliminated variable $x_{k}$ would be expressed as follows:

    \begin{equation}
        S'=
        \begin{cases}
            x_{k} \geqslant - \frac{1}{w_{k,1}}[(\sum_{i \in [\![1,N]\!] \setminus \lbrace{k}\rbrace}  w_{i,1}x_{i}) + b_{1}]
            \\  x_{k} \geqslant  - \frac{1}{w_{k,2}}[(\sum_{i \in [\![1,N]\!] \setminus \lbrace{k}\rbrace}  w_{i,2}x_{i}) + b_{2}]
            \\ \vdots 
            \\  x_{k} \geqslant - \frac{1}{w_{k,P}}[(\sum_{i \in [\![1,N]\!] \setminus \lbrace{k}\rbrace}  w_{i,P}x_{i}) + b_{P}]
            \\x_{k} \leqslant - \frac{1}{w_{k,P+1}}[(\sum_{i \in [\![1,N]\!] \setminus \lbrace{k}\rbrace}  w_{i,P+1}x_{i}) + b_{P+1}]
            \\  x_{k} \leqslant  - \frac{1}{w_{k,P+2}}[(\sum_{i \in [\![1,N]\!] \setminus \lbrace{k}\rbrace}  w_{i,P+2}x_{i}) + b_{P+2}]
            \\ \vdots 
            \\  x_{k} \leqslant - \frac{1}{w_{k,P+Q}}[(\sum_{i \in [\![1,N]\!] \setminus \lbrace{k}\rbrace}  w_{i,P+Q}x_{i}) + b_{P+Q}]
        \end{cases}
        \end{equation}

        \begin{equation*}
        \Rightarrow 
        \begin{cases}
         x_{k} \geqslant max_{j \in [\![1,P]\!]}(- \frac{1}{w_{k,j}}[(\sum_{i \in [\![1,N]\!] \setminus \lbrace{k}\rbrace}  w_{i,j}x_{i}) + b_{j}]))
        \\ x_{k} \leqslant min_{j \in [\![P+1,P+Q]\!]}(- \frac{1}{w_{k,j}}[(\sum_{i \in [\![1,N]\!] \setminus \lbrace{k}\rbrace}  w_{i,j}x_{i}) + b_{j}]))
        \end{cases} 
        \end{equation*}

        The next step is to keep $(12)$ consistent. If consistent solutions exist, they will be able to be found by processing the same Fourier-Motzkin 
        elimination process on the newly introduce inequalities system $S'$. According to $(12)$, inequalities to satisfy can be rewritten as :

        \begin{equation*}
             S'=
            \begin{cases}
                - \frac{1}{w_{k,P+1}}[(\sum_{i \in [\![1,N]\!] \setminus \lbrace{k}\rbrace}  w_{i,P+1}x_{i}) + b_{P+1}] 
                \geqslant - \frac{1}{w_{k,1}}[(\sum_{i \in [\![1,N]\!] \setminus \lbrace{k}\rbrace}  w_{i,1}x_{i}) + b_{1}]
                \\ \vdots 
                \\ - \frac{1}{w_{k,P+Q}}[(\sum_{i \in [\![1,N]\!] \setminus \lbrace{k}\rbrace}  w_{i,P+Q}x_{i}) + b_{P+Q}] 
                \geqslant - \frac{1}{w_{k,1}}[(\sum_{i \in [\![1,N]\!] \setminus \lbrace{k}\rbrace}  w_{i,1}x_{i}) + b_{1}]
                \\- \frac{1}{w_{k,P+1}}[(\sum_{i \in [\![1,N]\!] \setminus \lbrace{k}\rbrace}  w_{i,P+1}x_{i}) + b_{P+1}] 
                \geqslant - \frac{1}{w_{k,2}}[(\sum_{i \in [\![1,N]\!] \setminus \lbrace{k}\rbrace}  w_{i,2}x_{i}) + b_{2}]
                \\ \vdots 
                \\ - \frac{1}{w_{k,P+Q}}[(\sum_{i \in [\![1,N]\!] \setminus \lbrace{k}\rbrace}  w_{i,P+Q}x_{i}) + b_{P+Q}] 
                \geqslant - \frac{1}{w_{k,2}}[(\sum_{i \in [\![1,N]\!] \setminus \lbrace{k}\rbrace}  w_{i,2}x_{i}) + b_{2}]
                \\ \vdots 
                \\ \vdots 
                \\- \frac{1}{w_{k,P+1}}[(\sum_{i \in [\![1,N]\!] \setminus \lbrace{k}\rbrace}  w_{i,P+1}x_{i}) + b_{P+1}] 
                \geqslant - \frac{1}{w_{k,P}}[(\sum_{i \in [\![1,N]\!] \setminus \lbrace{k}\rbrace}  w_{i,P}x_{i}) + b_{P}]
                \\ \vdots 
                \\ - \frac{1}{w_{k,P+Q}}[(\sum_{i \in [\![1,N]\!] \setminus \lbrace{k}\rbrace}  w_{i,P+Q}x_{i}) + b_{P+Q}] 
                \geqslant - \frac{1}{w_{k,P}}[(\sum_{i \in [\![1,N]\!] \setminus \lbrace{k}\rbrace}  w_{i,P}x_{i}) + b_{P}]

            \end{cases}
            \end{equation*}

        \begin{equation*}
            \Rightarrow  S'=
            \begin{cases}
                - \frac{1}{w_{k,P+1}}[(\sum_{i \in [\![1,N]\!] \setminus \lbrace{k}\rbrace}  w_{i,P+1}x_{i}) + b_{P+1}] 
                + \frac{1}{w_{k,1}}[(\sum_{i \in [\![1,N]\!] \setminus \lbrace{k}\rbrace}  w_{i,1}x_{i}) + b_{1}] \geqslant 0
                \\ \vdots 
                \\ - \frac{1}{w_{k,P+Q}}[(\sum_{i \in [\![1,N]\!] \setminus \lbrace{k}\rbrace}  w_{i,P+Q}x_{i}) + b_{P+Q}] 
                 + \frac{1}{w_{k,1}}[(\sum_{i \in [\![1,N]\!] \setminus \lbrace{k}\rbrace}  w_{i,1}x_{i}) + b_{1}] \geqslant 0
                \\- \frac{1}{w_{k,P+1}}[(\sum_{i \in [\![1,N]\!] \setminus \lbrace{k}\rbrace}  w_{i,P+1}x_{i}) + b_{P+1}] 
                + \frac{1}{w_{k,2}}[(\sum_{i \in [\![1,N]\!] \setminus \lbrace{k}\rbrace}  w_{i,2}x_{i}) + b_{2}] \geqslant 0
                \\ \vdots 
                \\ - \frac{1}{w_{k,P+Q}}[(\sum_{i \in [\![1,N]\!] \setminus \lbrace{k}\rbrace}  w_{i,P+Q}x_{i}) + b_{P+Q}] 
                + \frac{1}{w_{k,2}}[(\sum_{i \in [\![1,N]\!] \setminus \lbrace{k}\rbrace}  w_{i,2}x_{i}) + b_{2}] \geqslant 0
                \\ \vdots 
                \\ \vdots 
                \\- \frac{1}{w_{k,P+1}}[(\sum_{i \in [\![1,N]\!] \setminus \lbrace{k}\rbrace}  w_{i,P+1}x_{i}) + b_{P+1}] 
                + \frac{1}{w_{k,P}}[(\sum_{i \in [\![1,N]\!] \setminus \lbrace{k}\rbrace}  w_{i,P}x_{i}) + b_{P}] \geqslant 0
                \\ \vdots 
                \\ - \frac{1}{w_{k,P+Q}}[(\sum_{i \in [\![1,N]\!] \setminus \lbrace{k}\rbrace}  w_{i,P+Q}x_{i}) + b_{P+Q}] 
                + \frac{1}{w_{k,P}}[(\sum_{i \in [\![1,N]\!] \setminus \lbrace{k}\rbrace}  w_{i,P}x_{i}) + b_{P}] \geqslant 0
                
            \end{cases}
            \end{equation*}\\
            \
    Let $\{w'_{i,j}\}_{i \in [\![1,N]\!] \setminus \lbrace{k}\rbrace ,j \in [\![1,PQ]\!]} \in \mathbb{R}^{N+PQ-1},
    w'_{i,j} =  \frac{w_{i,\ceil*{j/Q}}}{w_{k,\ceil*{j/Q}}} - \frac{w_{i,P+1+(j-1[Q])}}{w_{k,P+1+(j-1[Q])}} $\\
    Let $\{b'_{j}\}_{j \in [\![1,PQ]\!]} \in \mathbb{R}^{PQ},  
    b'_{j} =  \frac{b_{\ceil*{j/Q}}}{w_{k,\ceil*{j/Q}}} - \frac{b_{P+1+(j-1[Q])}}{w_{k,P+1+(j-1[Q])}}$\\\\
    Then :

    \begin{equation*}
        S'=
        \begin{cases}
             (\sum_{i \in [\![1,N]\!] \setminus \lbrace{k}\rbrace} w'_{i,1}x_{i}) + b'_{1} \geqslant 0
            \\  (\sum_{i \in [\![1,N]\!] \setminus \lbrace{k}\rbrace} w'_{i,2}x_{i}) + b'_{2} \geqslant 0
            \\ \vdots 
            \\  (\sum_{i \in [\![1,N]\!] \setminus \lbrace{k}\rbrace} w'_{i,PQ}x_{i}) + b'_{PQ} \geqslant 0
        \end{cases}
        \end{equation*}
        As $S'$ has the same form as $S$, the same process can be applied to eliminate another free variable.\\\\

    \paragraph{Extrema Approach} \   
    \\
    After Fourier-Motzkin elimination, the linear inequalities system $S$ remains with same signs real coefficients such as :

    \begin{equation*}
        \forall (w_{i,j_1},w_{i,j_2}),i \in [\![1,N]\!],(j_1,j_2) \in [\![1,M]\!]^{2},w_{i,j_1}w_{i,j_2}>0 
    \end{equation*}

    As a reminder :\\\\
    $\forall i \in [\![1,N]\!],\forall j \in [\![1,M]\!],w_{i,j},b_{i} \in \mathbb{R}$

    \begin{equation*}
        S=
        \begin{cases}
             (\sum_{i=1}^{N} w_{i,1}x_{i}) + b_{1} \geqslant 0
            \\  (\sum_{i=1}^{N} w_{i,2}x_{i}) + b_{2} \geqslant 0
            \\ \vdots 
            \\  (\sum_{i=1}^{N} w_{i,M}x_{i}) + b_{M} \geqslant 0
        \end{cases}
        \end{equation*}
        \
        \\\\
        To find solutions of $S$ a similar process will be applied to it. In fact, it is a particular case of Fourier-Motzkin elimination where there is 
        no opposite sign to proceed as previously. To solve this particular case, first select a variable $x_{k} \in \textbf{x},k \in [\![1,N]\!]$.\\
        While there still remain unsolved inequalities, process as follows:\\\\
        $\forall j \in [\![1,M]\!],{w_{k,j}}>0,$ then :
        \\\\
        $(\sum_{i=1}^{N} w_{i,j}x_{i}) + b_{j} \geqslant 0 \Rightarrow 
        x_{k} \geqslant - \frac{1}{w_{k,j}}[(\sum_{i \in [\![1,N]\!] \setminus \lbrace{k}\rbrace}  w_{i,j}x_{i}) + b_{j}]
        $ 
        \\\\
        Otherwise if ${w_{k,j}}<0,$
        \\\\
        $(\sum_{i=1}^{N} w_{i,j}x_{i}) + b_{j} \geqslant 0 \Rightarrow 
        x_{k} \leqslant - \frac{1}{w_{k,j}}[(\sum_{i \in [\![1,N]\!] \setminus \lbrace{k}\rbrace}  w_{i,j}x_{i}) + b_{j}]
        $ 
        \\\\
        The variable $x_{k}$ can be eliminated from the system as it can be expressed as follows:
        
        \begin{equation}
            \begin{cases}
                x_{k} \geqslant - \frac{1}{w_{k,1}}[(\sum_{i \in [\![1,N]\!] \setminus \lbrace{k}\rbrace}  w_{i,1}x_{i}) + b_{1}]
                \\  x_{k} \geqslant  - \frac{1}{w_{k,2}}[(\sum_{i \in [\![1,N]\!] \setminus \lbrace{k}\rbrace}  w_{i,2}x_{i}) + b_{2}]
                \\ \vdots 
                \\  x_{k} \geqslant - \frac{1}{w_{k,n}}[(\sum_{i \in [\![1,N]\!] \setminus \lbrace{k}\rbrace}  w_{i,n}x_{i}) + b_{n}]
            \end{cases}
            \end{equation}
            \begin{equation*}
                \Rightarrow  x_{k} \geqslant max_{ j \in [\![1,M]\!]}(- \frac{1}{w_{k,j}}[(\sum_{i \in [\![1,N]\!] \setminus \lbrace{k}\rbrace}  w_{i,j}x_{i}) + b_{j}]))
        \end{equation*}
        \\
        Notice that the case where ${w_{k,j}}<0$ can be solved with the same method by reversing the sign of the previous inequalities and by replacing the $max$ function
        by a $min$ function.\\

        At this point, each variable left is free or expressed by trivial inequalities that can be reduced as a maximum (or minimum) evaluation.
        For example, let $v$ a remaining variable.
        \\\\
        Let $n,m \in \mathbb{N},\textbf{c} \in \mathbb{R}^{n+m}$
    
        \begin{equation}
            \begin{cases}
                 v \geqslant c_{1}
                \\  v \geqslant c_{2}
                \\ \vdots 
                \\  v \geqslant c_{n}
                \\  v \leqslant c_{n+1}
                \\ \vdots
                \\  v \leqslant c_{n+m}
            \end{cases}
            \Rightarrow 
            \begin{cases}
             v \geqslant max(c_{i},i \in [\![1,n]\!]) 
            \\ v \leqslant min(c_{j},j \in [\![n+1,n+m]\!])
            \end{cases}
            \end{equation}

            As $\textbf{c}$ is a vector of $n$ known real values, evaluating $v$ can be done in $O(n)$ time complexity.
            Notice that if inequalities of $(14)$ are inconsistent, the system $S$ has no solutions. 
            Because every variables are expressed as system of inequalities of known symbol, solutions of $S$ 
            can be found by iteratively computing the maximum (or minimum) value for each of them.\\
            The previous algorithm gives a convenient way to solve linear inequalities as it ends in an exponential time complexity.
            It however needs more computation (in polynomial time) to enable its solutions to be exploitable.\\
            \\\

    \end{appendices}

\end{document}